\documentclass{svproc}

\usepackage{times}
\usepackage{epsfig}
\usepackage{graphicx}
\usepackage{amsmath}
\usepackage{amssymb}
\usepackage{color}
\usepackage{subfigure}
\usepackage{epstopdf}
\usepackage{setspace}
\usepackage{url}
\usepackage{cite}

\makeatletter
\newcommand*\bigcdot{\mathpalette\bigcdot@{.5}}
\newcommand*\bigcdot@[2]{\mathbin{\vcenter{\hbox{\scalebox{#2}{$\m@th#1\bullet$}}}}}
\makeatother

\usepackage{url}

\begin{document}
\mainmatter              
\title{DAC: Deep Autoencoder-based Clustering, a General Deep Learning Framework of Representation Learning}
\titlerunning{Deep Autoencoder-based Clustering}  
%
\author{Si Lu\inst{1} \and Ruisi Li}

\authorrunning{Si Lu et al.} 
\tocauthor{Si Lu and Ruisi Li}
\institute{Portland State University}

\maketitle              

\begin{abstract}
Clustering performs an essential role in many real world applications, such as market research, pattern recognition, data analysis, and image processing. However, due to the high dimensionality of the input feature values, the data being fed to clustering algorithms usually contains noise and thus could lead to in-accurate clustering results. While traditional dimension reduction and feature selection algorithms could be used to address this problem, the simple heuristic rules used in those algorithms are based on some particular assumptions. When those assumptions does not hold, these algorithms then might not work. In this paper, we propose DAC, Deep Autoencoder-based Clustering, a generalized data-driven framework to learn clustering representations using deep neuron networks. Experiment results show that our approach could effectively boost performance of the K-Means clustering algorithm on a variety types of datasets.
\keywords{clustering, K-Means, representation learning, deep neuron networks, deep autoencoder}
\end{abstract}
\section{Introduction}

Clustering is the task of grouping samples such that the ones in the same group are more similar to each other than to the ones in other groups. Nowadays, clustering performs as a basic and essential pre-processing step of many real world applications. For example, it could be used to help with fake news identification\cite{hosseinimotlagh2018unsupervised}, document analysis\cite{zhao2002evaluation}, marketing and sales, etc. Specifically, clustering algorithms can figure out useful information for the applications via grouping according to a variety of data similarity metrics and data grouping schemes. For example, similar patches could be used for image denoising \cite{bm3d, buades:nlm,chen:external} or depth enhancement \cite{lu2014depth}, and clustering could be used to find good similar patches \cite{lu2019good}.

To let the samples be properly assigned to different groups(called clusters), meaningful feature values of the samples need to be obtained first. However, in real world applications, the data we get is often of high dimensions\cite{han2011data} and usually contains noise, making the clustering difficult to succeed. For example, in the MNIST dataset\cite{lecun1998gradient}, each input hand-written digit image has 784 pixels. While we know some pixels (e.g. the ones at image corners) might not be as useful as others(e.g. the ones around image centers), it is difficult to manually distinguish them in clustering.

Traditional dimensionality reduction algorithms, namely Principle Component Analysis(PCA)\cite{pearson1901liii}, Linear Discriminant Analysis(LDA)\cite{duda2001pattern}, and Canonical Correlation Analysis(CCA)\cite{sun2005new},  could be used to reduce the number of features. In addition, feature selection algorithms can be used to select from the original feature values a set of useful and noiseless ones. These algorithms aim to extract the core information given the redundant and correlated input high-dimension data features. However, these algorithms often fail mainly due to two reasons. Firstly, most of them require complex mathematical analysis, which is difficult and time consuming as well. Secondly, their is no single approach that could work for all types of datasets. Different datasets could have different dimensions, data sizes and even might be used in totally different applications. Some datasets are linear and some of them are non-linear. As a result, it is difficult to find a way to generally work on all types of datasets. 

Recently, due to the emerging of the powerful deep neuron networks, deep learning-based approaches have been introduced to learn better data representations and achieve appealing performance improvements for clustering algorithms. 
One simple approach is to learn representations using deep auto-encoders. Specifically, the original input high dimensional features are fed into a encoder that generates a low dimensional output. This output is further fed to a decoder that tries to recover the raw input data as much as possible. However, most of the existing approaches\cite{pu2016variational,Yang_2019_CVPR} are using images as input and thus using convolutional neuron networks in their work. 

In this paper, we propose Deep Clustering Autoencoder, a simple but more general framework for representation learning that takes feature vectors as input. Thus, our approach could be applied to more generalized datasets. In addition, according to the group labels, we propose a scheme to adaptively weight all input features. We combine this estimated weight with the loss function computation during training. Experiment results show that our approach could effectively improve the performance of K-Means clustering algorithm on different types of datasets, namely MNIST, Fashion-MNIST\cite{xiao2017fashion}, as well as Human Activities and Postural Transitions Data Set (HAPT)\cite{reyes2016transition}. 

The rest of the paper is organized as follows: in section 2 we describe the overview of our deep autoencoder-based clustering. We then describe the deep autoencoder for representation learning in more details in section 3. We finally show experimental results in section 4 and conclude in section 5.

\begin{figure}[htb]
\vspace{2.5cm}
\begin{center}
\includegraphics[width=1.0\textwidth]{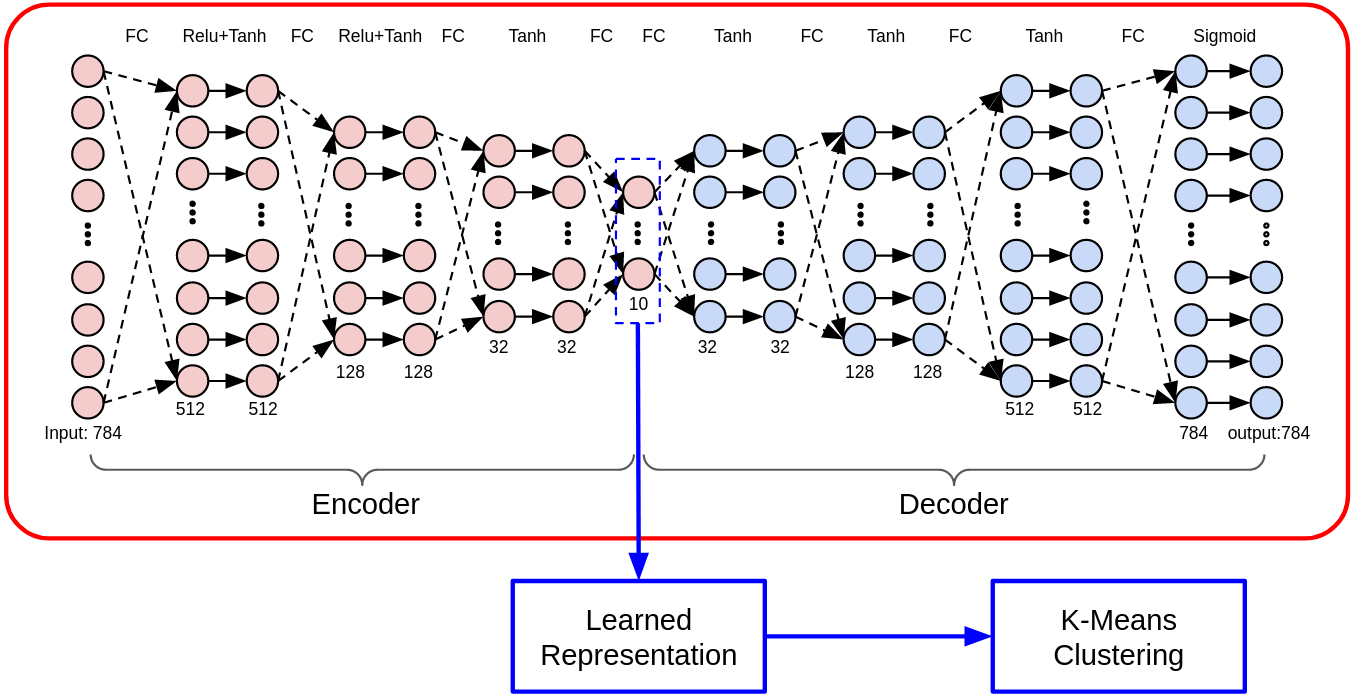} \\
\caption{\footnotesize{Overview of our Deep Autoencoder-based Clustering on MNSIT dataset. The autoencoder (consists of an encoder and a decoder) tries to encode and decode the input features such that the decoded output is as close to the input as possible. The input size is 28X28 = 784, the size of the learned low-dimension representation is 10. In the testing stage, the learned encoder output is then fed into the classic K-Means algorithm to do clustering. }
\label{fig:overview}}
\end{center}
\end{figure}

\section{Overview of Deep Autoencoder-based Clustering}

Figure \ref{fig:overview} shows an overview of our deep autoencoder-based clustering framework. There are two main steps: training and clustering testing. In the training step, a deep autoencoder with an encoder and a decoder is trained using the training set. Here a flattened input vector is fed into the multilayer deep encoder which has a low dimensional learned representation. This learned representation is further fed into a decoder that tries to recover an output of the same size as the input. The training process of this autoencoder tries to reconstruct the input as much as possible. In the following clustering step, we apply the autoencoder to the testing set. The output of the encoder (learned representations) is then fed to a classic K-Means algorithm to do clustering. The learned low dimensional representation vector contains key information of the given input, and thus yield better clustering results.

\section{Deep Autoencoder for Representation Learning}
\label{seq:method}
The architecture of our deep autoencoder for representation learning is shown in Figure \ref{fig:overview}. As could be seen, the model is not as complex as some of the advanced neuron networks. The reason is that we do not want our model to over-fit in two-folds. First, we do not want our model to over-fit on the training dataset over the testing dataset. Second, we do not want our model to over-fit on the reconstruction problem it-self over the clustering problem. Thus, we select a model of reasonable median complexity.

\subsection{Encoder}
The encoder aims to encode or compress the input data into a smaller size representation, and at the same time preserve as much key information as possible. As shown in Figure \ref{fig:overview}, the encoder consists of 8 layers, include the input layer and the learned representation output layer. Here the input layer is being normalized such that all its values is in the range of $(0,1)$. Specifically, from the beginning, each larger layer is fully connected to the next smaller layer followed by a couple of activation layers. There are mainly two types of activation layers, Relu and Tanh, as shown in Equation \ref{eq:Relu} and \ref{eq:Tanh}. Adding the Relu layers could introduce non-linearity to our model, making it more robust against non-linear input data. The Tanh layer, on the other hand, could transform the data into a normalized range of $(-1,1)$, to alleviate the gradient vanishing/exploding problem.

\begin{equation} \label{eq:Relu}
Relu(x) = max(0,x)
\end{equation}

\begin{equation} \label{eq:Tanh}
\begin{split}
\cosh(x) &={\frac{{\rm e}^x+{\rm e}^{-x}}{2}} \\
\sinh(x) &={\frac{{\rm e}^x-{\rm e}^{-x}}{2}} \\
\tanh(x) &= \frac{\sinh(x)}{\cosh(x)}=\frac{{\rm e}^x-{\rm e}^{-x}}{{\rm e}^x+{\rm e}^{-x}}
\end{split}
\end{equation}

\subsection{Decoder}
The decoder aims to decode or decompress the encoded output to reconstruct the original input data as much as possible. It contains nine layers, include the input layer, which is the output of the encoder, and the final output layer. Specifically, each smaller layer is fully connected to the next larger layer followed by a Tanh activation layer. In addition, the decoder has a Sigmoid activation layer (shown in Equation \ref{eq:Sigmoid}) at the final stage to enforce the output values lie into the range of $(0,1)$. 

\begin{equation} \label{eq:Sigmoid}
Sigmoid(x) = \frac{1}{1+e^{-x}}
\end{equation}

\subsection{Objective Function}

\subsubsection{Clustering-weighted MSE Loss}
While the goal of the classic autoencoder is to reconstruct the original input as much as possible, it counts each input feature value equally. However, it is possible that each individual input feature contributes differently to the final clustering results. For example, in MNIST dataset, the pixels at the four corners of almost all images are of the same color black (with zero intensity input values), thus have no impact to the final clustering at all. On the other hand, some pixels around the center of the images are likely to perform more important roles. We thus propose a scheme to compute a clustering-weighted MSE loss to let the autoencoder focus more on the reconstruction of more important input feature values, as shown in Equation \ref{eq:LCMSE}. 

\begin{equation} \label{eq:LCMSE}
L_{cmse} = \frac{\sum_{i=1}^{n}w_i(y_i-\hat{y_i})^2}{n}
\end{equation}

Here $w_i$ is the weight of each feature. It is computed using all ith feature values sampled from a subset of the training dataset with $m$ samples. Denote all ith feature values as $\{x_ik|k=1,2,..,m\}$ and the corresponding ground truth group/cluster labels of the $m$ samples $\{l_k|k=1,2,..,m\}$. The corresponding feature weight will be large if both of the two following conditions are met. First, all sampled values in the same groups/clusters have small differences. Second, all sampled values in different groups/clusters have large differences. Thus, the weight is computed as:

\begin{equation} \label{eq:weight}
w_{i} = \frac{\sum\limits_{l_p=l_q}e^{-(x_{ip}-x_{iq})^2}}{\sum\limits_{l_p=l_q}1} \bigcdot \frac{\sum\limits_{l_p \neq l_q}(1-e^{-(x_{ip}-x_{iq})^2)}}{\sum\limits_{l_p \neq l_q}1}
\end{equation}

\begin{figure}[htb]
\begin{center}
\includegraphics[width=0.5\textwidth]{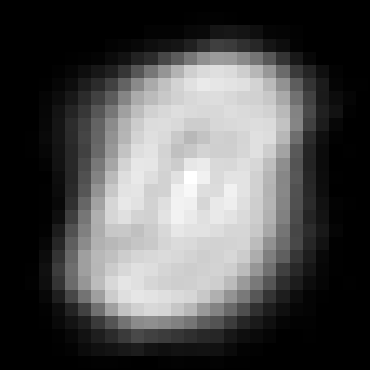} \\
\caption{\footnotesize{A map of the clustering weight computed for MNIST dataset using 1000 samples from the training set. It could be seen that pixels at boundaries and corners are less important than the ones around image centers.}
\label{fig:clustering_weight}}
\end{center}
\end{figure}

Figure \ref{fig:clustering_weight} shows a map of the clustering weight computed for MNIST dataset using 1000 samples from the training set. Pixels at boundaries and corners are less important than the ones around image centers, thus have smaller weights(white means larger weights).

\subsubsection{Final Objective Function}

The final objective function then combines the Clustering-weighted MSE Loss and a standard L2 norm regularization, as shown in Equation \ref{eq:FinalLoss}. Here the L2 norm regularization $L_r$ is computed using all parameters from the autoencoder. $\beta$ is a balancing factor with a default value of $0.00001$.

\begin{equation} \label{eq:FinalLoss}
L = L_{cmse} + \beta \dot L_{r}
\end{equation}

\section{Experimental Results}
\subsection{Dataset}
\begin{figure}[htb]
\begin{center}
\includegraphics[width=1.0\textwidth]{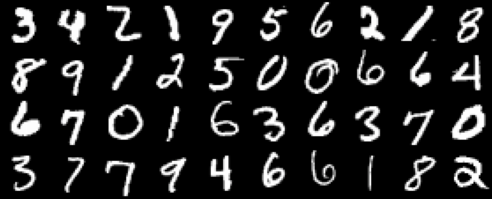} \\
\caption{\footnotesize{Samples of the MNIST dataset.}
\label{fig:mnistSample}}
\end{center}
\end{figure}

We evaluate our approach on the classic MNIST hand-written digits dataset. This dataset has $50,000$ images as the training set and $10,000$ images as the testing set. There are 10 groups in total. We show some samples of MNIST dataset in Figure \ref{fig:mnistSample}.

\subsection{Measurement Metrics}
To evaluate our framework, we apply our trained encoder to the testing dataset. We then compare the generated representations from our trained encoder to the raw input features by applying them to the K-Means algorithm. To measure the performance of clustering algorithms, we use the Adjusted Rand Index (ARI). Specifically, this metrics computes a similarity between two clustering results by considering all pairs of samples and counting pairs that are assigned in the same or different clusters in the predicted and ground truth clustering results. The proposed approach is denoted as DAC.

\subsection{Experiment setup}
We implement our framework in Python and PyTorch and test it on a desktop with RTX 2080-Ti. We train the autoencoder for 200 epochs using Adam Optimization Algorithm. The initial learning rate is set to 0.003 and will decrease with the number of epochs during training.

\subsection{Results on MNIST}

\begin{figure}[htb]
\begin{center}
\includegraphics[width=1.0\textwidth]{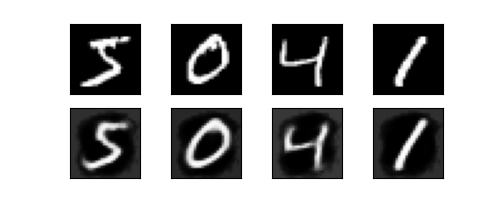} \\
\caption{\footnotesize{Sample results of our trained autoencoder on MNIST dataset. Top: raw input images. Bottom: reconstructed images}
\label{fig:mnistAutoEncoderSample}}
\end{center}
\end{figure}

Table \ref{tab:res_mnist} shows the quantitative performance of the proposed approach in terms of $ARI$. Comparing to the raw K-Means algorithm, our approach ($DAC$) boosts the K-Means algorithm's performace from $0.3477$ to $0.6624$, which is a $90.50\%$ boost. We also show some of the reconstructed results by our trained autoencoder in Figure \ref{fig:mnistAutoEncoderSample}. It shows that our trained autoencoder can properly reconstruct the raw input hand-written digits.

\begin{table}[htb]
\centering
\caption {\footnotesize{Clustering results on MNIST testing dataset. \label{tab:res_mnist}}}
\begin{tabular}{ccc}
  \hline
  & K-Mesn& DAC \\
  \hline
  $ARI$ & 0.3477  & 0.6624  \\
  \hline
\end{tabular}
\end{table}

\subsection{Results on other datasets}

To test the robustness of our approach against different data types, we apply our method to two other datasets: Fashion-MNIST\cite{xiao2017fashion}, and Human Activities and Postural Transitions Data Set (HAPT)\cite{reyes2016transition}.

\begin{figure}[htb]
\begin{center}
\includegraphics[width=1.0\textwidth]{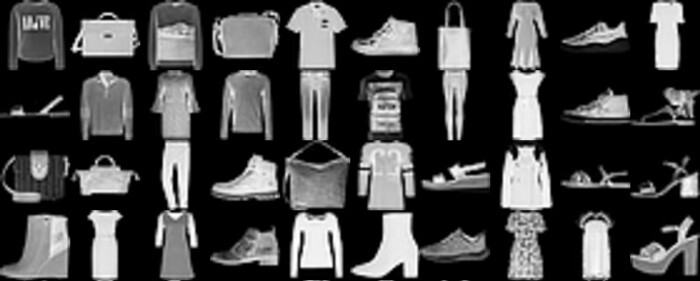} \\
\caption{\footnotesize{Samples of the Fashion-MNIST dataset.}
\label{fig:fmnistSample}}
\end{center}
\end{figure}

Fashion-MNIST is a similar dataset to MNIST, with the same image format and image size. It has $60,000$ images as training set and $10,000$ images as testing set. The only difference is the content: it contains images of 10 types of clothes. The ten categories are shown in Table \ref{tab:fmnist_labels}. We show some samples of this dataset in Figure \ref{fig:fmnistSample}.

\begin{table}[htb]
\centering
\caption {\footnotesize{Fashion-MNIST category labels. \label{tab:fmnist_labels}}}
\begin{tabular}{p{20mm}p{20mm}p{20mm}p{20mm}p{20mm} }
  \hline
  T-shirt/top & Trouser & Pullover & Dress & Coat\\
  \hline
  Sandal      & Shirt   & Sneaker  & Bag   & Ankle boot \\
  \hline
\end{tabular}
\end{table}

Human Activities and Postural Transitions Data Set is a dataset that has been captured by smart phone's sensors \cite{reyes2016transition}. The authors captured 3-axial linear acceleration and 3-axial angular velocity at a constant rate of 50Hz using the embedded accelerometer and gyroscope of the device, which is a smartphone (Samsung Galaxy S II). There are 30 volunteers whose ages are in the range of 19-48 years old. In their data capturing experiment, the volunteers was doing one of twelve activities. There are six basic activities: three static postures (standing, sitting, lying) and three dynamic activities (walking, walking downstairs and walking upstairs). Another six postural transitions that occurred between the static postures have also been added to the dataset. These are: stand-to-sit, sit-to-stand, sit-to-lie, lie-to-sit, stand-to-lie, and lie-to-stand. All twelve types of activities are shown in Table \ref{tab:hapt_labels}.

\begin{table}[htb]
\centering
\caption {\footnotesize{HAPT category labels. \label{tab:hapt_labels}}}
\begin{tabular}{p{30mm}p{30mm}p{30mm}p{30mm} }
  \hline
  walking  & walking upstairs & walking downstairs & sitting \\
  \hline
  standing & laying & stand to sit & sit to stand \\
  \hline
  standing & laying & stand to sit & sit to stand \\
  \hline
  sit to lie & lie to sit & stand to lie  & lie to stand \\
  \hline
\end{tabular}
\end{table}

The sensor signals (accelerometer and gyroscope) were then denoised by some noise filters. The authors then sampled in fixed-width sliding windows of 2.56 sec and $50\%$ overlap (128 readings/window), leading to a sample size of 561 features. Each sample is captured when the volunteer is doing one type of activities. During the capture process, $70\%$ of the volunteers were randomly selected to generate the training set and $30\%$ were selected to generate the testing set. In total, this dataset has 7767 samples for training and 3162 samples for testing.

\begin{figure}[tb]
\begin{center}
\includegraphics[width=1.0\textwidth]{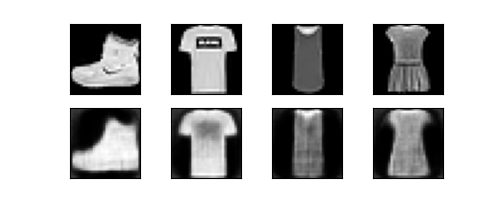} \\
\caption{\footnotesize{Sample results of our trained on Fashion-MNIST dataset. Top: raw input images. Bottom: reconstructed images}
\label{fig:fmnistAutoEncoderSample}}
\end{center}
\end{figure}

\begin{figure}[htb]
\vspace{2.5cm}
\begin{center}
\includegraphics[width=1.0\textwidth]{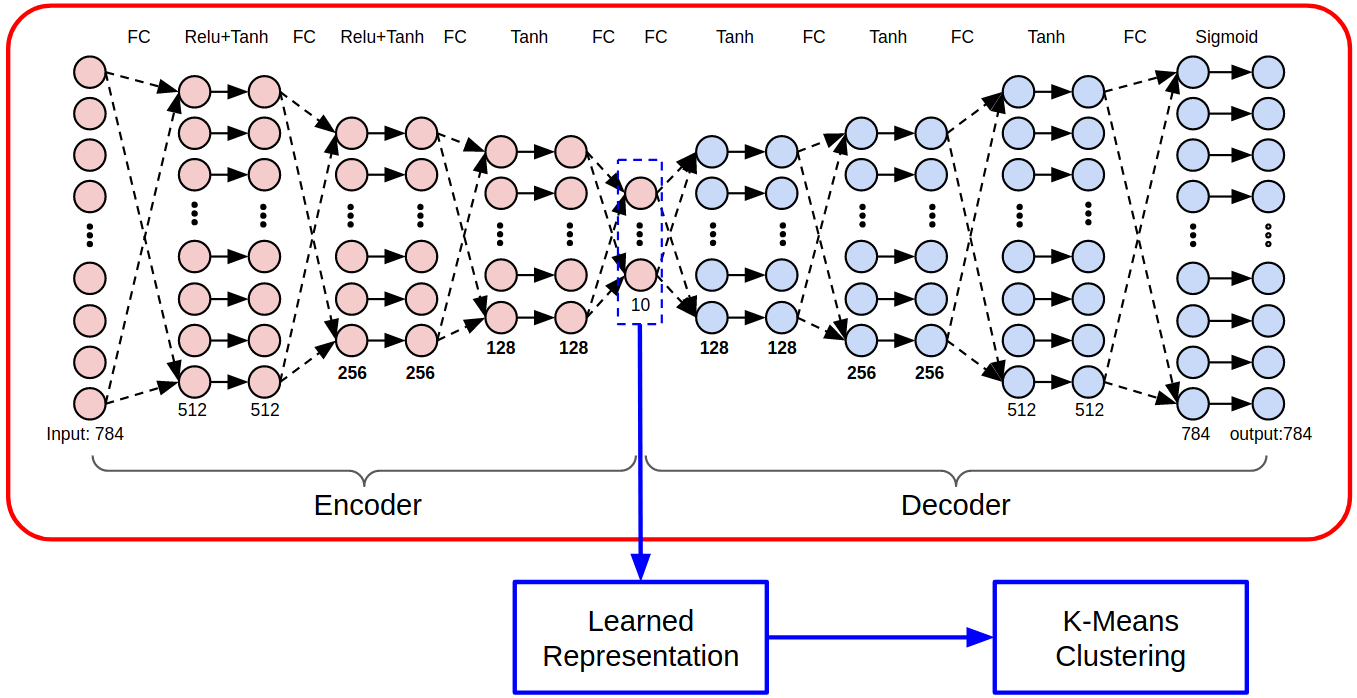} \\
\caption{\footnotesize{Overview of our Deep Autoencoder-based Clustering on Fashion-MNSIT dataset.}
\label{fig:fmnist_net}}
\vspace{0.5cm}
\includegraphics[width=1.0\textwidth]{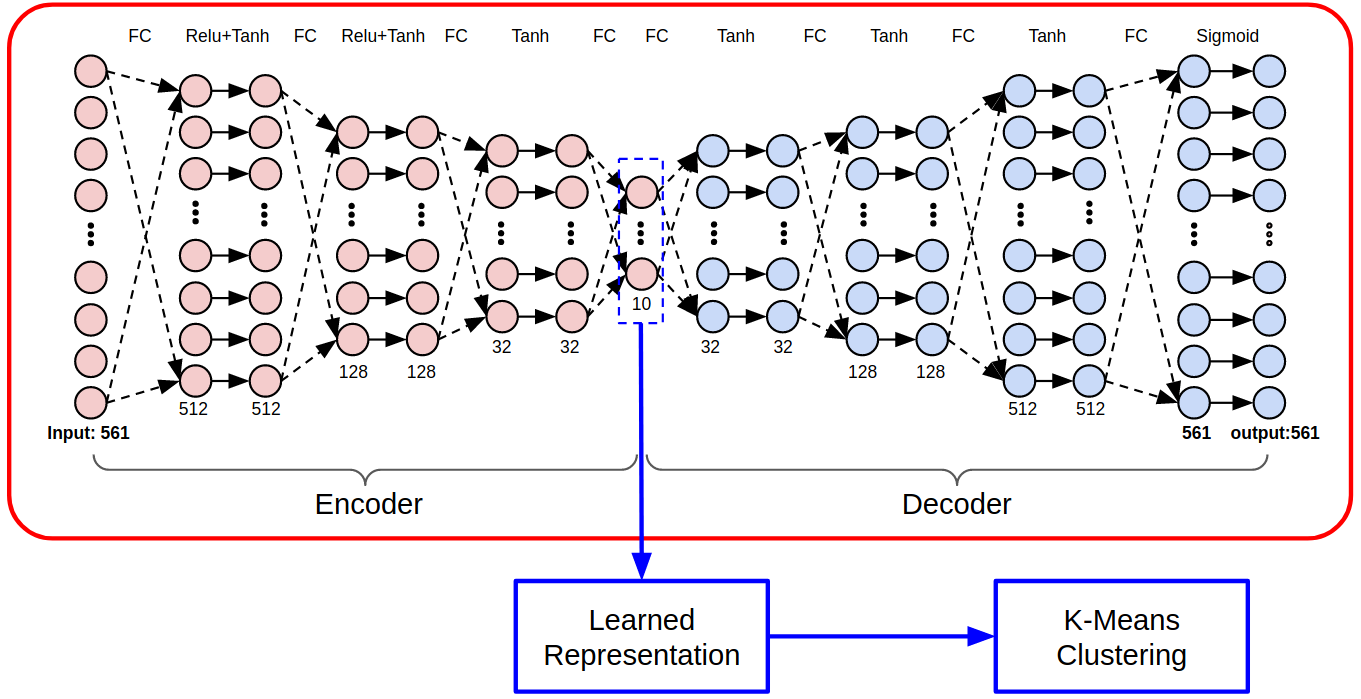} \\
\caption{\footnotesize{Overview of our Deep Autoencoder-based Clustering on HAPT dataset.}
\label{fig:hapt_net}}
\end{center}
\end{figure}

We apply our method to Fashion-MNIST dataset and report the results in Table \ref{tab:res_fmnist}. Here as the Fashion-MNIST is a more complex dataset, we modified the autoencoder and show the modified autoencoder architecture in Figure \ref{fig:fmnist_net}. It can be seen that comparing to using raw input features in K-Means clustering, our method boosts ARI from 0.3039 to 0.4702, yields to a improvement of $54.7\%$. 

We then apply our method to the HAPT dataset and report the results in Table \ref{tab:hapt_labels}. Here as this dataset's inputs are of lower dimension than MNIST, we modified the autoencoder accordingly and show the modified autoencoder architecture in Figure \ref{fig:hapt_net}. It can be seen that even with this temporal sequence dataset, our method could effectively improve the K-Means algorithm's performance by $30\%$. These results also show that our method could be generally applied to other data types. We also show some of the reconstructed results by our trained autoencoder in Figure \ref{fig:fmnistAutoEncoderSample}. It shows that our trained autoencoder can properly reconstruct the raw input fashion images.

\begin{table}[htb]
\centering
\caption {\footnotesize{Clustering results on Fashion-MNIST testing dataset. \label{tab:res_fmnist}}}
\begin{tabular}{ccc}
  \hline
  & K-Mesn& DAC \\
  \hline
  $ARI$ & 0.3039  & 0.4702  \\
  \hline
\end{tabular}
\end{table}

\begin{table}[htb]
\centering
\caption {\footnotesize{Clustering results on HAPT testing dataset. \label{tab:res_hapt}}}
\begin{tabular}{ccc}
  \hline
  & K-Mesn& DAC \\
  \hline
  $ARI$ & 0.4290  & 0.5594  \\
  \hline
\end{tabular}
\end{table}

\section{Conclusion}

In this paper, we propose DAC, Deep Autoencoder-based Clustering, a generalized data-driven framework to learn low dimensional clustering representations using trained deep neuron networks. Specifically, we train a multi-layer deep autoencoder to encode and decode the raw input samples. The encoded output of the encoder is then fed to a classic K-Means algorithm to do clustering. We design a scheme to compute a clustering-based weight in the training objective function to train the autoencoder and let it focus more on the reconstruction of more important features. Experimental results show that our approach could effectively boost the performance of a classic clustering algorithm: K-Means by $30\%$ to $90\%$ on MNIST dataset. In addtion, our method could be also applied to other types of clustering datasets, such as Fashion-MNIST and Human Activities and Postural Transitions Data Set (HAPT). Experimental results show that our framework could still be able to improve K-Means algorithm's performance by as much as $55\%$

\clearpage

\bibliographystyle{spmpsci}
\bibliography{template.bib}

\begin{thebibliography}{10}
\providecommand{\url}[1]{{#1}}
\providecommand{\urlprefix}{URL }
\expandafter\ifx\csname urlstyle\endcsname\relax
  \providecommand{\doi}[1]{DOI~\discretionary{}{}{}#1}\else
  \providecommand{\doi}{DOI~\discretionary{}{}{}\begingroup
  \urlstyle{rm}\Url}\fi

\bibitem{buades:nlm}
Buades, A., Coll, B., Morel, J.: A non-local algorithm for image denoising.
\newblock In: IEEE Conference on Computer Vision and Pattern Recognition
  (CVPR), vol.~2, pp. 60--65 (2005)

\bibitem{chen:external}
Chen, F., Zhang, L., Yu, H.: External patch prior guided internal clustering
  for image denoising.
\newblock In: IEEE International Conference on Computer Vision (ICCV), pp.
  603--611 (2015)

\bibitem{bm3d}
Dabov, K., Foi, A., Katkovnik, V., Egiazarian, K.: Image denoising by sparse
  3-d transform-domain collaborative filtering.
\newblock IEEE Transactions on Image Processing \textbf{16}(8), 2080--2095
  (2007)

\bibitem{duda2001pattern}
Duda, R.O., Hart, P.E., Stork, D.G.: Pattern classification. john wiley \&
  sons.
\newblock Inc., New York \textbf{2} (2001)

\bibitem{han2011data}
Han, J., Pei, J., Kamber, M.: Data mining: concepts and techniques.
\newblock Elsevier (2011)

\bibitem{hosseinimotlagh2018unsupervised}
Hosseinimotlagh, S., Papalexakis, E.E.: Unsupervised content-based
  identification of fake news articles with tensor decomposition ensembles.
\newblock In: Proceedings of the Workshop on Misinformation and Misbehavior
  Mining on the Web (MIS2) (2018)

\bibitem{lecun1998gradient}
LeCun, Y., Bottou, L., Bengio, Y., Haffner, P.: Gradient-based learning applied
  to document recognition.
\newblock Proceedings of the IEEE \textbf{86}(11), 2278--2324 (1998)

\bibitem{lu2019good}
Lu, S.: Good similar patches for image denoising.
\newblock In: 2019 IEEE Winter Conference on Applications of Computer Vision
  (WACV), pp. 1886--1895. IEEE (2019)

\bibitem{lu2014depth}
Lu, S., Ren, X., Liu, F.: Depth enhancement via low-rank matrix completion.
\newblock In: Proceedings of the IEEE conference on computer vision and pattern
  recognition, pp. 3390--3397 (2014)

\bibitem{pearson1901liii}
Pearson, K.: Liii. on lines and planes of closest fit to systems of points in
  space.
\newblock The London, Edinburgh, and Dublin Philosophical Magazine and Journal
  of Science \textbf{2}(11), 559--572 (1901)

\bibitem{pu2016variational}
Pu, Y., Gan, Z., Henao, R., Yuan, X., Li, C., Stevens, A., Carin, L.:
  Variational autoencoder for deep learning of images, labels and captions.
\newblock Advances in neural information processing systems \textbf{29},
  2352--2360 (2016)

\bibitem{reyes2016transition}
Reyes-Ortiz, J.L., Oneto, L., Sam{\`a}, A., Parra, X., Anguita, D.:
  Transition-aware human activity recognition using smartphones.
\newblock Neurocomputing \textbf{171}, 754--767 (2016)

\bibitem{sun2005new}
Sun, Q.S., Zeng, S.G., Liu, Y., Heng, P.A., Xia, D.S.: A new method of feature
  fusion and its application in image recognition.
\newblock Pattern Recognition \textbf{38}(12), 2437--2448 (2005)

\bibitem{xiao2017fashion}
Xiao, H., Rasul, K., Vollgraf, R.: Fashion-mnist: a novel image dataset for
  benchmarking machine learning algorithms.
\newblock arXiv preprint arXiv:1708.07747  (2017)

\bibitem{Yang_2019_CVPR}
Yang, X., Deng, C., Zheng, F., Yan, J., Liu, W.: Deep spectral clustering using
  dual autoencoder network.
\newblock In: Proceedings of the IEEE/CVF Conference on Computer Vision and
  Pattern Recognition (CVPR) (2019)

\bibitem{zhao2002evaluation}
Zhao, Y., Karypis, G.: Evaluation of hierarchical clustering algorithms for
  document datasets.
\newblock In: Proceedings of the eleventh international conference on
  Information and knowledge management, pp. 515--524 (2002)

\end{thebibliography}

\end{document}